\documentclass[runningheads]{llncs}
\usepackage{graphicx}

\usepackage{tikz}
\usepackage{comment}
\usepackage{mathtools}
\usepackage{amsmath,amssymb} 
\usepackage{color}

\usepackage{booktabs}
\usepackage{pifont}
\usepackage{multirow}
\usepackage{array}
\usepackage{wrapfig}
\usepackage{diagbox}

\usepackage{listings}
\usepackage{algorithm}

\usepackage[accsupp]{axessibility}  

\usepackage[pagebackref=true,breaklinks=true,letterpaper=true,colorlinks,bookmarks=false]{hyperref}

\usepackage{orcidlink}

\newcommand{\parsection}[1]{\noindent\textbf{#1:}~}

\usepackage{etoolbox}
\makeatletter
\AfterEndEnvironment{algorithm}{\let\@algcomment\relax}
\AtEndEnvironment{algorithm}{\kern2pt\hrule\relax\vskip-3pt\@algcomment}
\let\@algcomment\relax
\newcommand\algcomment[1]{\def\@algcomment{\footnotesize#1}}
\renewcommand\fs@ruled{\def\@fs@cfont{\bfseries}\let\@fs@capt\floatc@ruled
	\def\@fs@pre{\hrule height.8pt depth0pt \kern2pt}%
	\def\@fs@post{}%
	\def\@fs@mid{\kern2pt\hrule\kern2pt}%
	\let\@fs@iftopcapt\iftrue}
\makeatother

\begin{document}
\pagestyle{headings}
\mainmatter
\def\ECCVSubNumber{1212}  

\title{CMD: Self-supervised 3D Action Representation Learning with Cross-modal Mutual Distillation} 

\titlerunning{CMD}

\author{
Yunyao Mao\inst{1}\orcidlink{0000-0002-9427-9086} \and
Wengang Zhou\inst{1,2,*}\orcidlink{0000-0003-1690-9836}\and
Zhenbo Lu\inst{2}\orcidlink{0000-0002-0918-7524}\and \\
Jiajun Deng\inst{1}\orcidlink{0000-0001-9624-7451}\and
Houqiang Li\inst{1,2,*}\orcidlink{0000-0003-2188-3028}
}
\authorrunning{Y. Mao et al.}
%
\institute{
CAS Key Laboratory of Technology in GIPAS, EEIS Department,\\University of Science and Technology of China
\and
Institute of Artificial Intelligence, Hefei Comprehensive National Science Center \\
\email{myy2016@mail.ustc.edu.cn},~~\email{zhwg@ustc.edu.cn},~~\email{luzhenbo@iai.ustc.edu.cn},\\
\email{dengjj@ustc.edu.cn},~~\email{lihq@ustc.edu.cn}
}
\maketitle

\begin{abstract}
In 3D action recognition, there exists rich complementary information between skeleton modalities.
Nevertheless, how to model and utilize this information remains a challenging problem for self-supervised 3D action representation learning.
In this work, we formulate the cross-modal interaction as a bidirectional knowledge distillation problem.
Different from classic distillation solutions that transfer the knowledge of a fixed and pre-trained teacher to the student, in this work, the knowledge is continuously updated and bidirectionally distilled between modalities.
To this end, we propose a new \textbf{C}ross-modal \textbf{M}utual \textbf{D}istillation (CMD) framework with the following designs.
On the one hand, the neighboring similarity distribution is introduced to model the knowledge learned in each modality, where the relational information is naturally suitable for the contrastive frameworks. 
On the other hand, asymmetrical configurations are used for teacher and student to stabilize the distillation process and to transfer high-confidence information between modalities.
By derivation, we find that the cross-modal positive mining in previous works can be regarded as a degenerated version of our CMD.
We perform extensive experiments on NTU RGB+D 60, NTU RGB+D 120, and PKU-MMD II datasets. Our approach outperforms existing self-supervised methods and sets a series of new records. The code is available at: \url{https://github.com/maoyunyao/CMD}
\keywords{Self-supervised 3D action recognition, contrastive learning}
\end{abstract}

\makeatletter
\def\blfootnote{\xdef\@thefnmark{}\@footnotetext}
\makeatother

\blfootnote{* Corresponding authors: Wengang Zhou and Houqiang Li}

\section{Introduction}

Human action recognition, one of the fundamental problems in computer vision, has a wide range of applications in many downstream tasks, such as behavior analysis, human-machine interaction, virtual reality, \emph{etc}. 
Recently, with the advancement of human pose estimation algorithms \cite{openpose,fang2017rmpe,xu2020deep}, skeleton-based 3D human action recognition has attracted increasing attention for its light-weight and background-robust characteristics. However, fully-supervised 3D action recognition \cite{Chen_2021_ICCV,du2015hierarchical,ke2017new,li2019actional,Li_2021_ICCV,liu2020disentangling,shi2019skeleton,Shi_2021_ICCV,si2019attention,zhang2019view,zhang2020semantics,zhang2020context} requires large amounts of well-annotated skeleton data for training, which is rather labor-intensive to acquire. 
In this paper, we focus on the self-supervised settings, aiming to avoid the laborious workload of manual annotation for 3D action representation learning.

\begin{figure}[t!]
	\centering
	\includegraphics[width=1.0\linewidth]{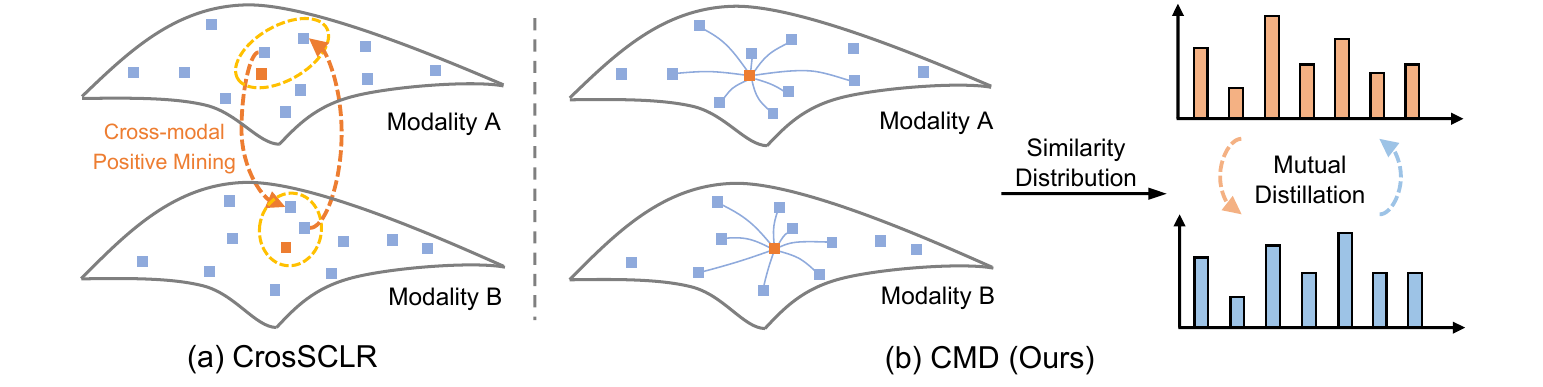}
	\caption{CrosSCLR \cite{li20213d} vs. CMD (Ours). Given handful negative samples, CrosSCLR performs cross-modal positive mining according to  the cosine similarity between embeddings. The nearest neighbor of the positive query in modality A will serve as an additional positive sample in modality B, and vice versa. In our approach, we reformulate cross-modal interaction as a bidirectional knowledge distillation problem, with similarity distribution that models the modality-specific knowledge.}\label{fig:overview}
\end{figure}

To learn robust and discriminative representation, many celebrated pretexts like motion prediction, jigsaw puzzle recognition, and masked reconstruction have been extensively studied in early works \cite{lin_ms2l_mm,misra2016shuffle,nie2020unsupervised,noroozi2016unsupervised,su2020predict,zheng2018unsupervised}. 
Recently, the contrastive learning frameworks \cite{chen2020simple,he2020momentum,van2018representation} have been introduced to the self-supervised 3D action recognition community \cite{lin_ms2l_mm,rao2021augmented}. It achieves great success thanks to the capability of learning discriminative high-level semantic features. 
However, there still exist unsolved problems when applying contrastive learning on skeletons. 
On the one hand, the success of contrastive learning heavily relies on performing data augmentation \cite{chen2020simple}, but the skeletons from different videos are unanimously considered as negative samples. Given the limited action categories, it would be unreasonable to just ignore potential similar instances, since they may belong to the same category as the positive one. On the other hand, cross-modal interactive learning is largely overlooked in early contrastive learning-based attempts \cite{lin_ms2l_mm,rao2021augmented}, yet integrating multimodal information \cite{cheng2020skeleton,SSVOD,transvgpp,liang2019three,2sagcn2019cvpr,9234715} is the key to improving the performance of 3D action recognition.

To tackle these problems, CrosSCLR \cite{li20213d} turns to cross-modal positive mining (see Figure~\ref{fig:overview} (a)) and sample reweighting. Though effective, it suffers the following limitations. Firstly, the positive sample mining requires reliable preliminary knowledge, thus the representation in each modality needs to be optimized independently in advance, leading to a sophisticated two-stage training process. Secondly, the contrastive context, defined as the similarity between the positive query and negative embeddings, is treated as individual weights of samples in complementary modalities to participate in the optimization process. Such implicit knowledge exchange lacks a holistic grasp of the rich contextual information. Besides, the cross-modal consistency is also not explicitly guaranteed.

In this work, we go beyond heuristic positive sample mining and reformulate cross-modal interaction as a general bidirectional knowledge distillation \cite{hinton2015distilling} problem. As shown in Figure~\ref{fig:overview} (b), in the proposed Cross-modal Mutual Distillation (CMD) framework, the neighboring similarity distribution is first extracted in each modality. It describes the relationship of the sample embedding with respect to its nearest neighbors in the customized feature space. Compared with individual features \cite{hinton2015distilling} or logits \cite{romero2014fitnets}, such relational information is naturally suitable for modeling the knowledge learned with contrastive frameworks. Based on the relational information, bidirectional knowledge distillation between each two modalities is performed via explicit cross-modal consistency constraints. Since the representation in each skeleton modality is trained from scratch and there is no intuitive teacher-student relationship between modalities, embeddings from the momentum updated key encoder along with a smaller temperature are used for knowledge modeling on the teacher side, so as to stabilize the distillation process and highlight the high-confidence information in each modality.

Compared to previous works, the advantages of our approach are three-fold:
\textbf{i)} Instead of heuristically reweighting training samples, the contextual information in contrastive learning is treated as a whole to model the modality-specific knowledge, explicitly ensuring the cross-modal consistency during distillation. 
\textbf{ii)} Unlike cross-modal positive sample mining, our approach does not heavily rely on the initial representation, thus is free of the sophisticated two-stage training. This largely benefits from the probabilistic knowledge modeling strategy. Moreover, the positive mining is also mathematically proved to be a special case of the proposed cross-modal distillation mechanism under extreme settings.
\textbf{iii)} The proposed CMD is carefully designed to be well integrated into the existing contrastive framework with almost no extra computational overhead introduced.

We perform extensive experiments on three prevalent benchmark datasets: NTU RGB+D 60 \cite{shahroudy2016ntu}, NTU RGB+D 120 \cite{liu2020ntu}, and PKU-MMD II \cite{liu2017pku}. Our approach achieves state-of-the-art results on all of them under all evaluation protocols. It’s worth noting that the proposed cross-modal mutual distillation is easily implemented in a few lines of code. We hope this simple yet effective approach will serve as a strong baseline for future research.

\section{Related Work}
\parsection{Self-supervised Representation Learning}
Self-supervised learning methods can be roughly divided into two categories: generative and contrastive \cite{liu2021self}. Generative methods \cite{ballard1987modular,he2022masked,van2017neural} try to reconstruct the original input to learn meaningful latent representation. Contrastive learning \cite{chen2020simple,he2020momentum,van2018representation} aims to learn feature representation via instance discrimination. It pulls positive pairs closer and pushes negative pairs away. Since no labels are available during self-supervised contrastive learning, two different augmented versions of the same sample are treated as a positive pair, and samples from different instances are considered to be negative. In MoCo \cite{he2020momentum} and MoCo v2 \cite{chen2020improved}, the negative samples are taken from previous batches and stored in a queue-based memory bank. In contrast, SimCLR \cite{chen2020simple} and MoCo v3 \cite{chen2021empirical} rely on a larger batch size to provide sufficient negative samples. Similar to the contrastive context in~\cite{li20213d}, the neighboring similarity in this paper is defined as the normalized product between positive embedding and its neighboring anchors. Our goal is to transfer such modality-specific information between skeleton modalities to facilitate better contrastive 3D action representation learning.

\parsection{Self-supervised 3D Action Recognition}
Many previous works have been proposed to perform self-supervised 3D action representation learning. In LongT GAN \cite{zheng2018unsupervised}, an autoencoder-based model along with an additional adversarial training strategy are proposed. Following the generative paradigm, it learns latent representation via sequential reconstruction. Similarly, P\&C \cite{su2020predict} trains an encoder-decoder network to both predict and cluster skeleton sequences. To learn features that are more robust and separable, the authors also propose strategies to weaken the decoder, laying more burdens on the encoder. Different from previously mentioned methods that merely adopt a single reconstruction task, MS$^2$L \cite{lin_ms2l_mm} integrates multiple pretext tasks to learn better representation. In recent attempts \cite{li20213d,rao2021augmented,thoker2021skeleton,guo2022aimclr}, momentum encoder-based contrastive learning is introduced and better performance is achieved. Among them, CrosSCLR \cite{li20213d} is the first to perform cross-modal knowledge mining. It finds potential positives and re-weights training samples with the contrastive contexts from different skeleton modalities. However, the positive mining performed in CrosSCLR requires reliable initial representation, two-stage training is indispensable. Differently, in this paper, a more general knowledge distillation mechanism is introduced to perform cross-modal information interaction. Besides, the positive mining performed in CrosSCLR can be regarded as a special case of our approach.

\parsection{Similarity-based Knowledge Distillation}
Pairwise similarity has been shown to be useful information in relational knowledge distillation \cite{park2019relational,peng2019correlation,tung2019similarity}. In PKT \cite{passalis2018learning}, CompRess \cite{abbasi2020compress}, and SEED \cite{fang2021seed}, similarities of each sample with respect to a set of anchors are converted into a probability distribution, which models the structural information of the data. 
After that, knowledge distillation is performed by training the student to mimic the probability distribution of the teacher. 
Recently, contextual similarity information has also shown great potential in image retrieval \cite{ouyang2021contextual,wu2022contextual} and representation learning \cite{ALBEF,tejankar2021isd}. 
Our approach is partially inspired by these works. Differently, the cross-modal mutual distillation in our approach is designed
to answer the question of \textit{how to transfer the biased knowledge between complementary modalities during 3D action pre-training}.

\section{Method}

\begin{figure}[t!]
	\centering
	\includegraphics[width=1.0\linewidth]{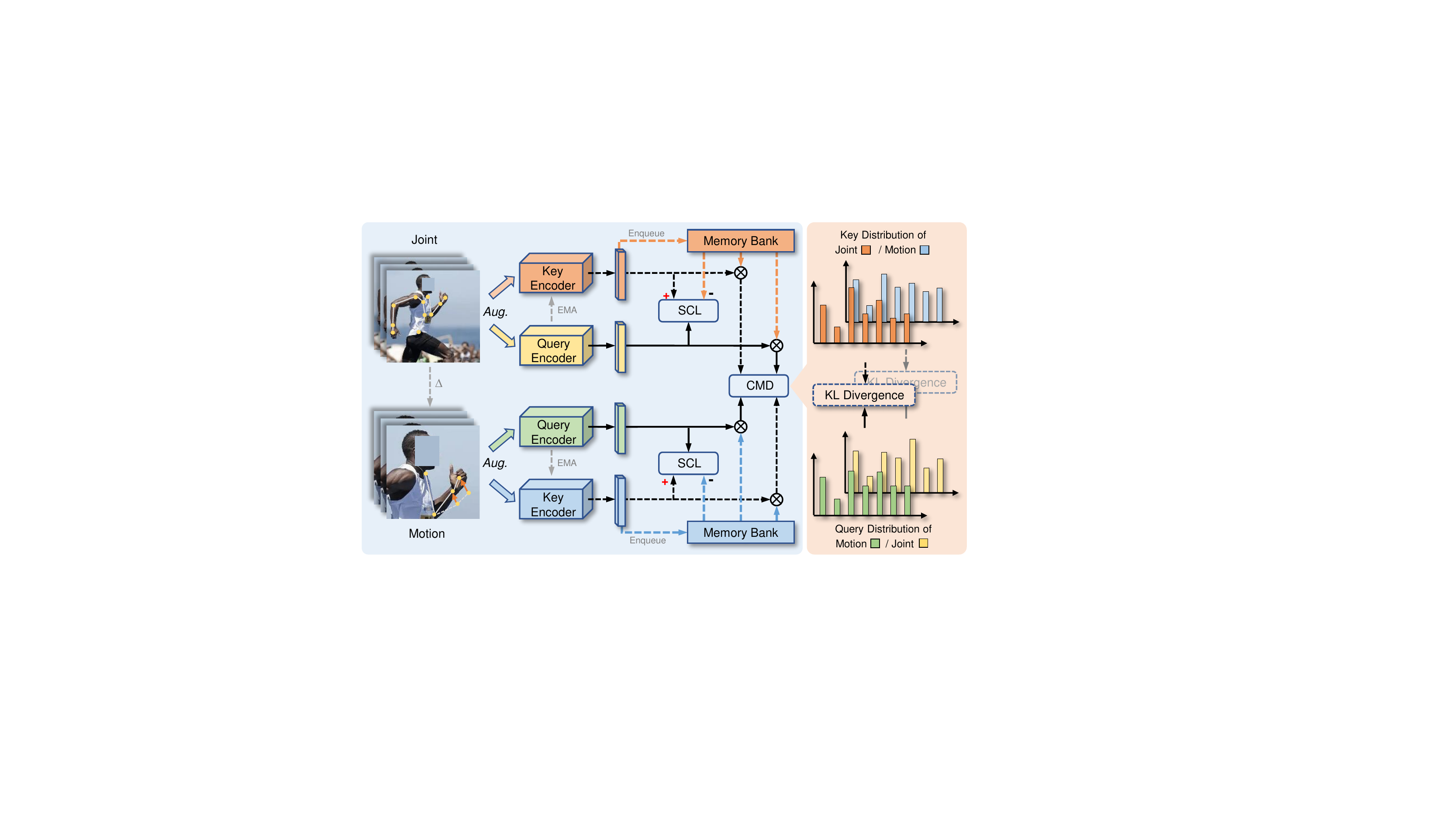}
	\caption{The overall pipeline of the proposed framework. It contains two modules, Single-modal Contrastive Learning (SCL) and Cross-modal Mutual Distillation (CMD). Given multiple skeleton modalities (\emph{e.g.} joint and motion) as input, the SCL module performs self-supervised contrastive learning in each modality and the CMD module simultaneously transfers the learned knowledge between modalities. SCL and CMD work collaboratively so that each modality learns more comprehensive representation.}
	\label{fig:arch}
\end{figure}

\subsection{Framework Overview}
\label{sec:overview}

By consolidating the idea of leveraging complementary information from cross-modal inputs to improve 3D action representation learning, we design the Cross-modal Mutual Distillation (CMD) framework. As shown in Figure~\ref{fig:arch}, the proposed CMD consists of two key components: Single-modal Contrastive Learning (SCL) and Cross-modal Mutual Distillation (CMD). Given multiple skeleton modalities (\emph{e.g.} joint, motion, and bone) as input, SCL is applied to each of them to learn customized 3D action representation. Meanwhile, in CMD, the knowledge learned by SCL is modeled by the neighboring similarity distributions, which describe the relationship between the sample embedding and its nearest neighbors. Cross-modal knowledge distillation is then performed by bidirectionally minimizing the KL divergence between the distributions corresponding to each modality. SCL and CMD run synchronously and cooperatively so that each modality learns more comprehensive representation.

\subsection{Single-modal Contrastive Learning}
\label{sec:single}
In this section, we revisit the single-modal contrastive learning as the preliminary of our approach, which has been widely adopted in many tasks like image/video recognition \cite {imagenet2009A,kuehne2011hmdb,soomro2012ucf101} and correspondence learning \cite{wang2021contrastive}. In self-supervised 3D action recognition, previous works like AS-CAL \cite{rao2021augmented},  CrosSCLR \cite{li20213d}, ISC \cite{thoker2021skeleton}, and AimCLR \cite{guo2022aimclr} also take the contrastive method MoCo v2 \cite{chen2020improved} as their baseline.

Given a single-modal skeleton sequence $x$, we first perform data augmentation to obtain two different views $x_q$ and $x_k$ (query and key). Then, two encoders are adopted to map the positive pair $x_q$ and $x_k$ into feature embeddings $z_q = E_q(x_q, \theta_q)$ and $z_k = E_k(x_k, \theta_k)$, where $E_q$ and $E_k$ denote query encoder and key encoder, respectively. $\theta_q$ and $\theta_k$ are the learnable parameters of the two encoders. Note that in MoCo v2, the key encoder is not trained by gradient descent but the momentum updated version of the query encoder:
$\theta_k \gets \alpha \theta_k + (1-\alpha)\theta_q$,
where $\alpha$ is a momentum coefficient that controls the updating speed.
During self-supervised pre-training, the noise contrastive estimation loss InfoNCE \cite{van2018representation} is used to perform instance discrimination, which is computed as follows:
\begin{equation}
	\begin{aligned}
		\mathcal{L}_{\text{SCL}}=
		-\log \frac{\exp (z_q ^\top z_k / \tau_c)}{\exp (z_q ^\top z_k / \tau_c)+\sum_{i=1}^{N} \exp \left(z_q ^\top m_{i} / \tau_c \right)},
	\end{aligned}
\end{equation}
where $\tau_c$ is a temperature hyper-parameter \cite{hinton2015distilling} that scales the distribution of instances and $m_i$ is the key embedding of negative sample. $N$ is the size of a queue-based memory bank $\mathbf{M}$ where all the negative key embeddings are stored. 
After the training of the current mini-batch, $z_k$ is enqueued as a new negative key embedding and the oldest embeddings in the memory bank are dequeued.

Under the supervision of the InfoNCE loss, the encoder is forced to learn representation that is invariant to data augmentations, thereby focusing on semantic information shared between positive pairs. 
Nevertheless, the learned representation is often modally biased, making it difficult to account for all data characteristics. 
Though it can be alleviated by test-time ensembling, several times the running overhead will be introduced. 
Moreover, the inherent limitations of the learned representation in each modality still exist. Therefore, during self-supervised pre-training, cross-modal interaction is essential.

\subsection{Cross-modal Mutual Distillation}
\label{sec:cmd}
While SCL is performed within each skeleton modality, the proposed CMD models the learned knowledge and transfers it between modalities. This enables each modality to receive knowledge from other perspectives, thereby alleviating the modal bias of the learned representation. Based on MoCo v2, CMD can be easily implemented in a few lines of code, as shown in Alg.~\ref{alg:code}.

\parsection{Knowledge Modeling} To perform knowledge distillation between modalities, we first need to model the knowledge learned in each modality in a proper way. It needs to take advantage of the existing contrastive learning framework to avoid introducing excessive computational overhead. Moreover, since the distillation is performed cross-modally for self-supervised learned knowledge, conventional methods that rely on individual features/logits are no longer applicable. 

Inspired by recent relational knowledge distillation works \cite{park2019relational,peng2019correlation,tung2019similarity}, we utilize the pairwise relationship between samples for modality-specific knowledge modeling. Given an embedding $z$ and a set of anchors $\{n_i\}_{i = 1,2,\cdots, K}$, we compute the similarities between them as $\text{sim}(z, n_i) = z^\top n_i, i = 1,2, \cdots, K.$

In the MoCo v2 \cite{chen2020improved} framework, there are a handful of negative embeddings stored in the memory bank. We can easily obtain the required anchors without additional model inference. Note that if all the negative embeddings are used as anchors, the set $\{z^\top m_i\}_{i = 1,2, \cdots, N}$ is exactly the contrastive context defined in \cite{li20213d}. In our approach, we select the top $K$ nearest neighbors of $z$ as the anchors. The resulting pairwise similarities are further converted into probability distributions with a temperature hyper-parameter $\tau$ :
\begin{equation}
	\label{eq:distribution}
	\begin{aligned}
		p_i(z, \tau) = \frac{\exp(z ^\top n_i  / \tau )}{\sum_{j=1}^{K} \exp(z ^\top n_j / \tau )}, i = 1,2, \cdots, K.
	\end{aligned}
\end{equation}
The obtained $ \boldsymbol{p}(z, \tau) = \{p_i(z, \tau)\}_{i = 1,2,\cdots, K}$ describes the distribution characteristic around the embedding $z$ in the customized feature space of each modality.

\parsection{Knowledge Distillation} Based on the aforementioned probability distributions, an intuitive way to perform knowledge distillation would be to directly establish consistency constraints between skeleton modalities. Different from previous knowledge distillation approaches that transfer the knowledge of a fixed and well-trained teacher model to the student, in our approach, the knowledge is continuously updated during self-supervised pre-training and each modality acts as both student and teacher.

To this end, based on the contrastive framework, we make two customized designs in the proposed approach: \textbf{i)} Different embeddings are used for teacher and student. As shown in Figure~\ref{fig:arch}, in MoCo v2 \cite{chen2020improved}, two augmented views of the same sample are encoded into query $z_q$ and key $z_k$, respectively. In our approach, the key distribution obtained in one modality is used to guide the learning of query distribution in other modalities, so that knowledge is transferred accordingly. Specifically, for the key embedding $z_k^a$ from modality A and the query embedding $z_q^b$ from modality B, we select the top $K$ nearest neighbors of $z_k^a$ as anchors and compute the similarity distributions as $\boldsymbol{p}(z_q^b, \tau)$ and $\boldsymbol{p}(z_k^a, \tau)$ according to Eq.~\ref{eq:distribution}. Knowledge distillation from modality A to modality B is performed by minimizing the following KL divergence:

\begin{equation}
	\begin{aligned}
		\mathrm{KL}\big(\boldsymbol{p}(z_k^a, \tau) || \boldsymbol{p}(z_q^b, \tau)\big) = \sum_{i=1}^{K} p_i(z_k^a, \tau) \cdot \log \frac{ p_i(z_k^a, \tau)}{p_i(z_q^b, \tau)}.
	\end{aligned}
\end{equation}
Since the key encoder is not trained with gradient, the teacher is not affected during unidirectional knowledge distillation. Moreover, the momentum updated key encoder provides more stable knowledge for the student to learn. \textbf{ii)} Asymmetric temperatures $\tau_t$ and $\tau_s$ are employed for teacher and student, respectively. Considering that there is no intuitive teacher-student relationship between modalities, a smaller temperature is applied for the teacher in CMD to emphasize the high-confidence information, as discussed in \cite{Tejankar_2021_ICCV}.

Since the knowledge distillation works bidirectionally, given two modalities A and B, the loss function for CMD is formulated as follows:
\begin{equation}
	\label{eq:loss_cmd}
	\begin{aligned}
		\mathcal{L}_{\text{CMD}} = \mathrm{KL}\big(\boldsymbol{p}(z_k^a, \tau_t) || \boldsymbol{p}(z_q^b, \tau_s)\big) + \mathrm{KL}\big(\boldsymbol{p}(z_k^b, \tau_t) || \boldsymbol{p}(z_q^a, \tau_s)\big).
	\end{aligned}
\end{equation}
Note that Eq.~\ref{eq:loss_cmd} can be easily extended if more modalities are involved. The final loss function in our approach is the combination of $\mathcal{L}_{\text{SCL}}$ and $\mathcal{L}_{\text{CMD}}$:
\begin{equation}
	\label{eq:loss_final}
	\begin{aligned}
		\mathcal{L} = \mathcal{L}_{\text{SCL}}^a + \mathcal{L}_{\text{SCL}}^b + \mathcal{L}_{\text{CMD}},
	\end{aligned}
\end{equation}
where the superscripts $a$ and $b$ denote modality A and B, respectively.

\begin{algorithm}[t]
	\caption{\small Pseudocode of the CMD module in a PyTorch-like style. 
	}
	\label{alg:code}
	\definecolor{codeblue}{rgb}{0.25,0.5,0.5}
	\definecolor{codekw}{rgb}{0.85, 0.18, 0.50}
	\lstset{
		backgroundcolor=\color{white},
		basicstyle=\fontsize{7.3pt}{7.3pt}\ttfamily\selectfont,
		columns=fullflexible,
		breaklines=true,
		captionpos=b,
		numbers=left,
		xleftmargin=2.3em,
		commentstyle=\fontsize{7.3pt}{7.3pt}\color{codeblue},
		keywordstyle=\fontsize{7.3pt}{7.3pt}\color{codekw},
		escapechar=\&
	}
	\begin{lstlisting}[language=python]
# z_q_a, z_q_b, z_k_a, z_k_b: query/key embeddings in modality A/B (BxC)
# queue_a, queue_b: queue of N keys in modality A/B (CxN)
# tau_s, tau_t: temperatures for student/teacher (scalars)

l_a, lk_a = torch.mm(z_q_a, queue_a), torch.mm(z_k_a, queue_a)  # compute similarities
l_b, lk_b = torch.mm(z_q_b, queue_b), torch.mm(z_k_b, queue_b)

lk_a_topk, idx_a = torch.topk(lk_a, K, dim=-1)  # select top K nearest neighbors
lk_b_topk, idx_b = torch.topk(lk_b, K, dim=-1)

loss_cmd = loss_kld(torch.gather(l_b, -1, idx_a) / tau_s, lk_a_topk / tau_t) # A to B
          + loss_kld(torch.gather(l_a, -1, idx_b) / tau_s, lk_b_topk / tau_t) # B to A

def loss_kld(inputs, targets):
     inputs, targets =  F.log_softmax(inputs, dim=1),  F.softmax(targets, dim=1)
     return F.kl_div(inputs, targets, reduction='batchmean')
	\end{lstlisting}
\end{algorithm}

\subsection{Relationship with Positive Mining}
\label{sec:relation}
\parsection{Cross-modal Positive Mining} Cross-modal positive mining is the most important component in CrosSCLR \cite{li20213d}, where the most similar negative sample is selected to boost the positive sets for contrastive learning in complementary modalities. The contrastive loss for modality B is reformulated as:
\begin{equation}
	\label{eq:cpm}
	\begin{aligned}
		\mathcal{L}_{\text{CPM}}^b
		&=
		-\log \frac{\exp ({z_q^b} ^\top {z_k^b} / \tau_c)}
		{\exp ({z_q^b} ^\top {z_k^b} / \tau_c)+\sum_{i=1}^{N} \exp ({z_q^b} ^\top m_{i}^b / \tau_c )} \\
        &
		-\log \frac{\exp ({z_q^b} ^\top {m_u^b} / \tau_c)}
		{\exp ({z_q^b} ^\top {z_k^b} / \tau_c)+\sum_{i=1}^{N} \exp ({z_q^b} ^\top m_{i}^b / \tau_c )} \\
		&=
		\mathcal{L}_{\text{SCL}}^b
		-\log \frac{\exp ({z_q^b} ^\top {m_u^b} / \tau_c)}
		{\exp ({z_q^b} ^\top {z_k^b} / \tau_c)+\sum_{i=1}^{N} \exp ({z_q^b} ^\top m_{i}^b / \tau_c )},		
	\end{aligned}
\end{equation}
where $u$ is the index of most similar negative sample in modality A.

\parsection{CMD with $\tau_t=0$ and $K=N$}
Setting temperature $\tau_t = 0$ and $K=N$, the key distribution $\boldsymbol{p}(z_k^a, \tau_t)$ in Eq.~\ref{eq:loss_cmd} will be an one-hot vector with the only $1$ at index $u$, and thus the loss works on modality B will be like:
\begin{equation}
	\label{eq:relation}
	\begin{aligned}
		\mathcal{L}^{b}
		&= \mathcal{L}_{\text{SCL}}^b + \mathcal{L}_{\text{CMD}}^{b} \\
		&= \mathcal{L}_{\text{SCL}}^b + \sum_{i=1}^{N} p_i(z_k^a, 0) \cdot \log \frac{ p_i(z_k^a, 0)}{p_i(z_q^b, \tau_s)} \\
		&= \mathcal{L}_{\text{SCL}}^b + 1 \cdot\log \frac{1}{p_u(z_q^b, \tau_s)} \\
		&= \mathcal{L}_{\text{SCL}}^b - \log \frac{\exp({z_q^b} ^\top m_u^b  / \tau_s )}{\sum_{j=1}^{N} \exp({z_q^b} ^\top m_j^b / \tau_s )}.
	\end{aligned}
\end{equation}
We can find that the loss $\mathcal{L}_{\text{CMD}}^{b}$ is essentially doing contrastive learning in modality B with the positive sample mined by modality A. Compared with Eq.~\ref{eq:cpm}, the only difference is that when the mined $m_u^b$ is taken as the positive sample, the key embedding $z_k^b$ is excluded from the denominator. The same result holds for modality A.
Thus we draw a conclusion that the cross-modal positive mining performed in CrosSCLR \cite{li20213d} can be regarded as a special case of our approach with the temperature of teacher $\tau_t=0$ and the number of neighbors $K=N$.

\section{Experiments}

\subsection{Implementation Details}
\parsection{Network Architecture} In our approach, we adopt a 3-layer Bidirectional GRU (BiGRU) as the base-encoder, which has a hidden dimension of 1024. Before the encoder, we additionally add a Batch Normalization \cite{ioffe2015batch} layer to stabilize the training process. Each skeleton sequence is represented in a two-actor manner, where the second actor is set to zeros if only one actor exists. The sequences are further resized to a temporal length of 64 frames.

\parsection{Self-supervised Pre-training} During pre-training, we adopt MoCo v2 \cite{chen2020improved} to perform single-modal contrastive learning.
The temperature hyper-parameter in the InfoNCE \cite{van2018representation} loss is 0.07. In cross-modal mutual distillation, the temperatures for teacher and student are set to 0.05 and 0.1, respectively. The number of neighbors $K$ is set to 8192. The SGD optimizer is employed with a momentum of 0.9 and a weight decay of 0.0001. The batch size is set to 64 and the initial learning rate is 0.01. For NTU RGB+D 60 \cite{shahroudy2016ntu} and NTU RGB+D 120 \cite{liu2020ntu} datasets, the model is trained for 450 epochs, the learning rate is reduced to 0.001 after 350 epochs, and the size of the memory bank $N$ is 16384. For PKU-MMD II \cite{liu2017pku} dataset, the total epochs are increased to 1000, and the learning rate drops at epoch 800. We adopt the same skeleton augmentations as ISC \cite{thoker2021skeleton}.

\subsection{Datasets and Metrics}
\parsection{NTU RGB+D 60 \cite{shahroudy2016ntu}} NTU-RGB+D 60 (NTU-60) is a large-scale multi-modality action recognition dataset which is captured by three Kinect v2 cameras. It contains 60 action categories and 56,880 sequences. The actions are performed by 40 different subjects (actors). In this paper, we adopt its 3D skeleton data for experiments. Specifically, each human skeleton contains 25 body joints, and each joint is represented as 3D coordinates. Two evaluation protocols are recommended by the authors: cross-subject (x-sub) and cross-view (x-view). For x-sub, action sequences performed by half of the 40 subjects are used as training samples and the rest as test samples. For x-view, the training samples are captured by camera 2 and 3 and the test samples are from camera 1.

\parsection{NTU RGB+D 120 \cite{liu2020ntu}} Compared with NTU-60, NTU-RGB+D 120 (NTU-120) extends the action categories from 60 to 120, with 114,480 skeleton sequences in total. The number of subjects is also increased from 40 to 106. Moreover, a new evaluation protocol named cross-setup (x-set) is proposed as a substitute for x-view. Specifically, the sequences are divided into 32 different setups according to the camera distances and background, with half of the 32 setups (even-numbered) used for training and the rest for testing.

\parsection{PKU-MMD \cite{liu2017pku}} PKU-MMD is a new benchmark for multi-modality 3D human action detection. It can also be used for action recognition tasks \cite{lin_ms2l_mm}. PKU-MMD has two phases, where Phase II is extremely challenging since more noise is introduced by large view variation. 
In this work, we evaluate the proposed method on Phase II (PKU-II) under the widely used cross-subject evaluation protocol, with 5,332 skeleton sequences for training and 1,613 for testing.

\parsection{Evaluation Metrics} We report the top-1 accuracy for all datasets.

\subsection{Comparison with State-of-the-art Methods}
In the section, the learned representation is utilized for 3D action classification under a variety of evaluation protocols. We compare the results with previous state-of-the-art methods. Note that during evaluation, we only take single skeleton modality (joint) as input by default, which is consistent with previous arts \cite{lin_ms2l_mm,su2020predict,thoker2021skeleton}. Integrating multiple skeleton modalities for evaluation can significantly improve the performance, but it will also incur more time overhead.

\begin{table}[t!]
	\centering
		\caption{Performance comparison on NTU-60, NTU-120, and PKU-II in terms of the linear evaluation protocol. Our approach achieves state-of-the-art performance on all of them, both when taking single skeleton modality as input and when ensembling multiple modalities during evaluation. The prefix “3s-" denotes multi-modal ensembling.}\label{tab:sota_comp_linear}
			\begin{tabular*}{1.0\linewidth}{p{0.21\linewidth} p{0.24\linewidth} p{0.095\linewidth}<{\centering}p{0.1\linewidth}<{\centering}p{0.095\linewidth}<{\centering}p{0.095\linewidth}<{\centering}p{0.11\linewidth}<{\centering}}
				\toprule
				\multirow{2}{*}{Method}   		  & \multirow{2}{*}{Modality}& \multicolumn{2}{c}{\textbf{NTU-60}}	& \multicolumn{2}{c}{\textbf{NTU-120}}	& \textbf{PKU-II} \\
				\cmidrule(lr){3-4} \cmidrule(lr){5-6} \cmidrule(lr){7-7}
				&  						& x-sub          & x-view  			& x-sub         	& x-set       & x-sub\\
				\midrule
				LongT GAN \cite{zheng2018unsupervised}   & Joint only				& 39.1           & 48.1   			& -             & - 				& 26.0\\
				MS$^2$L \cite{lin_ms2l_mm}               & Joint only				& 52.6           & -      			& -             & - 				& 27.6\\
				P\&C \cite{su2020predict}		         & Joint only				& 50.7           & 76.3       		& 42.7          & 41.7 				& 25.5\\
				AS-CAL \cite{rao2021augmented} 			 & Joint only				& 58.5           & 64.8   			& 48.6          & 49.2 				& -\\
				SeBiReNet \cite{nie2020unsupervised}     & Joint only				& -              & 79.7				& -      		& - 				& -\\
				AimCLR \cite{guo2022aimclr}	             & Joint only				& 74.3           & 79.7   			& -             & - 				& -\\
				ISC \cite{thoker2021skeleton}		     & Joint only				& 76.3           & 85.2   			& 67.1          & 67.9 				& 36.0\\
				CrosSCLR-B 	                     & Joint only				& 77.3           & 85.1   			& 67.1          & 68.6 				& 41.9\\
				\textbf{CMD (Ours)}               & Joint only				& \textbf{79.8}  & \textbf{86.9}	& \textbf{70.3} & \textbf{71.5} 	& \textbf{43.0}\\
				\midrule
				3s-CrosSCLR \cite{li20213d}			     &Joint+Motion+Bone		& 77.8           & 83.4   			& 67.9          & 66.7				& 21.2\\
				3s-AimCLR \cite{guo2022aimclr}	         &Joint+Motion+Bone		& 78.9           & 83.8   			& 68.2          & 68.8 				& 39.5\\
				3s-CrosSCLR-B            &Joint+Motion+Bone		& 82.1  & 89.2	& 71.6 & 73.4 	& 51.0\\
				\textbf{3s-CMD (Ours)}            &Joint+Motion+Bone		& \textbf{84.1}  & \textbf{90.9}	& \textbf{74.7} & \textbf{76.1} 	& \textbf{52.6}\\
				\bottomrule
			\end{tabular*}
\end{table}

\parsection{Linear Evaluation Protocol} For linear evaluation protocol, we freeze the pre-trained encoder and add a learnable linear classifier after it. The classifier is trained on the corresponding training set for 80 epochs with a learning rate of 0.1 (reduced to 0.01 and 0.001 at epoch 50 and 70, respectively). We evaluate the proposed method on the NTU-60, NTU-120, and PKU-II datasets. As shown in Table~\ref{tab:sota_comp_linear}, we include the recently proposed CrosSCLR \cite{li20213d}, ISC \cite{thoker2021skeleton}, and AimCLR \cite{guo2022aimclr} for comparison. Our approach outperforms previous state-of-the-art methods by a considerable margin on all the three benchmarks. Note that ISC and the proposed CMD share the same BiGRU encoder, which is different from the ST-GCN \cite{yan2018spatial} encoder in CrosSCLR. For a fair comparison, we additionally train a variation of CrossSCLR with BiGRU as its base-encoder (denoted as CrosSCLR-B). We can find that our method still outperforms it on all the three datasets, which shows the superiority of the proposed cross-modal mutual distillation.

\parsection{KNN Evaluation Protocol} An alternative way to use the pre-trained encoder for action classification is to directly apply a K-Nearest Neighbor (KNN) classifier to the learned features of the training samples. Following \cite{su2020predict}, we assign each test sample to the most similar class where its nearest neighbor is in (\emph{i.e.} KNN with k=1). As shown in Table~\ref{tab:sota_comp_knn}, we perform experiments on the NTU-60 and NTU-120 benchmarks and compare the results with previous works. For both datasets, our approach exhibits the best performance, surpassing CrosSCLR-B \cite{li20213d} by 4.5\%\textasciitilde6\% in the more challenging cross-subject and cross-setup protocols.

\begin{table}[t!]
	\begin{minipage}{.52\linewidth}
		\centering
		\caption{Performance comparison on NTU-60 and NTU-120 in terms of the KNN evaluation protocol. The learned representation exhibits the best performance on both datasets. Surpassing previous state-of-the-art methods by a considerable margin.}\label{tab:sota_comp_knn}
		\begin{tabular*}{1.0\linewidth}{p{0.38\linewidth} p{0.135\linewidth}<{\centering}p{0.15\linewidth}<{\centering}p{0.135\linewidth}<{\centering}p{0.135\linewidth}<{\centering}}
			\toprule
			\multirow{2}{*}{Method}   		  & \multicolumn{2}{c}{\textbf{NTU-60}}	& \multicolumn{2}{c}{\textbf{NTU-120}}\\
			\cmidrule(lr){2-3} \cmidrule(lr){4-5}
			& x-sub          & x-view  			& x-sub         	& x-set\\
			\midrule
			LongT GAN \cite{zheng2018unsupervised}  & 39.1           & 48.1   			& 31.5             & 35.5 \\
			P\&C \cite{su2020predict}              & 50.7           & 76.3      			& 39.5             & 41.8 \\
			ISC \cite{thoker2021skeleton}		    & 62.5           & 82.6   			& 50.6          & 52.3 \\
			CrosSCLR-B  		    & 66.1           & 81.3   			& 52.5          & 54.9 \\
			\textbf{CMD (Ours)}               & \textbf{70.6}  & \textbf{85.4}	& \textbf{58.3} & \textbf{60.9} \\
			\bottomrule
		\end{tabular*}
	\end{minipage}
	~
	\begin{minipage}{.46\linewidth}
		\centering
		\caption{Performance comparison on PKU-II in terms of the transfer learning evaluation protocol. The source datasets are NTU-60 and NTU-120. The representation learned by our approach shows the best transferability.}\label{tab:sota_comp_transfer}
		\begin{tabular*}{1.0\linewidth}{p{0.44\linewidth} p{0.25\linewidth}<{\centering}p{0.26\linewidth}<{\centering}}
			\toprule
			\multirow{2}{*}{Method}   		  & \multicolumn{2}{c}{\textbf{To PKU-II}} \\
			\cmidrule(lr){2-3}
			 & NTU-60          & NTU-120 \\
			\midrule
			LongT GAN \cite{zheng2018unsupervised}	 & 44.8           & -\\
			MS$^2$L \cite{lin_ms2l_mm}     		     & 45.8           & -\\
			ISC \cite{thoker2021skeleton}    		 & 51.1           & 52.3\\
			CrosSCLR-B		    & 54.0           & 52.8 \\
			\textbf{CMD (Ours)}               & \textbf{56.0}  & \textbf{57.0}\\
			\bottomrule
		\end{tabular*}
	\end{minipage}
\end{table}

\parsection{Transfer Learning Evaluation Protocol} In transfer learning evaluation protocol, we examine the transferability of the learned representation. Specifically, we first utilize the proposed framework to pre-train the encoder on the source dataset. Then the pre-trained encoder along with a linear classifier are finetuned on the target dataset for 80 epochs with a learning rate of 0.01 (reduced to 0.001 at epoch 50). We select NTU-60 and NTU-120 as source datasets, and PKU-II as the target dataset. We compare the proposed approach with previous methods LongT GAN \cite{zheng2018unsupervised}, MS$^2$L \cite{lin_ms2l_mm}, and ISC \cite{thoker2021skeleton} under the cross-subject protocol. As shown in Table~\ref{tab:sota_comp_transfer}, our approach exhibits 
superior performance on the PKU-II dataset after large-scale pre-training, outperforming previous methods by a considerable margin. This indicates that the representation learned by our approach is more transferable.

\begin{table}[t!]
	\caption{Performance comparison on NTU-60 in terms of the semi-supervised evaluation protocol. We randomly select a portion of the labeled data to fine-tune the pre-trained encoder, and the average of five runs is reported as the final performance. Our approach exhibits the state-of-the-art results compared with previous methods.} \label{tab:sota_semi}
		\centering
		\begin{tabular*}{1.0\linewidth}{p{0.23\linewidth}p{0.088\linewidth}<{\centering}p{0.088\linewidth}<{\centering}p{0.088\linewidth}<{\centering}p{0.088\linewidth}<{\centering}p{0.088\linewidth}<{\centering}p{0.088\linewidth}<{\centering}p{0.088\linewidth}<{\centering}p{0.088\linewidth}<{\centering}}
			\toprule
			\multirow{3}{*}{Method} & \multicolumn{8}{c}{\textbf{NTU-60 }}\\
			\cmidrule(lr){2-9}
			& \multicolumn{4}{c}{\textbf{x-view}} & \multicolumn{4}{c}{\textbf{x-sub}} \\
			\cmidrule(lr){2-5} \cmidrule(lr){6-9}
			&   (1\%) &  (5\%) & (10\%) &  (20\%) & (1\%) &  (5\%) & (10\%) &  (20\%) \\
			\midrule
			LongT GAN \cite{zheng2018unsupervised}   &  - & - & - & - & 35.2 & - & 62.0 & - \\
			MS$^2$L \cite{lin_ms2l_mm}  &  - & - & - & - & 33.1 & - & 65.1 & -  \\
			ASSL \cite{si2020adversarial} &  - & 63.6 & 69.8 & 74.7 & - & 57.3 & 64.3 & 68.0 \\
			ISC \cite{thoker2021skeleton} &  38.1   &  65.7   &72.5   &78.2   &35.7   &59.6   &65.9   &70.8 \\
			CrosSCLR-B \cite{li20213d} &     49.8   &  70.6   &77.0   &81.9   &48.6   &67.7   &72.4   &76.1 \\
			
			\textbf{CMD (Ours)}   &  \textbf{53.0}  &  \textbf{75.3}  &\textbf{80.2}  &\textbf{84.3}  &\textbf{50.6}  &\textbf{71.0}   &\textbf{75.4}   &\textbf{78.7} \\
			\midrule
			3s-CrosSCLR \cite{li20213d} &  50.0 & - & 77.8 & - & 51.1 & - & 74.4 & - \\
			3s-Colorization \cite{yang2021skeleton} &  52.5 & - & 78.9 & - & 48.3 & - & 71.7 & - \\
			3s-AimCLR \cite{guo2022aimclr} &  54.3 & - & 81.6 & - & 54.8 & - & 78.2 & - \\
			\textbf{3s-CMD (Ours)}   &  \textbf{55.5}  &  \textbf{77.2}  &\textbf{82.4}  &  \textbf{86.6}  &\textbf{55.6}  &  \textbf{74.3}  &\textbf{79.0}  &  \textbf{81.8} \\
			\bottomrule
		\end{tabular*}
\end{table}

\parsection{Semi-supervised Evaluation Protocol} In semi-supervised classification, both labeled and unlabeled data are included during training. 
Its goal is to train a classifier with better performance than the one trained with only labeled samples. 
For a fair comparison, we adopt the same strategy as ISC \cite{thoker2021skeleton}. The pre-trained encoder is fine-tuned together with the post-attached linear classifier on a portion of the corresponding training set. 
We conduct experiments on the NTU-60 dataset. 
As shown in Table~\ref{tab:sota_semi}, we report the evaluation results when the proportion of supervised data is set to 1\%, 5\%, 10\%, and 20\%, respectively. 
Compared with previous methods LongT GAN \cite{zheng2018unsupervised}, MS$^2$L \cite{lin_ms2l_mm}, ASSL \cite{si2020adversarial}, and ISC \cite{thoker2021skeleton}, our algorithm exhibits superior performance. 
For example, with the same baseline, the proposed approach outperforms ISC and CrosSCLR-B by a large margin. 
We also take 3s-CrosSCLR \cite{li20213d}, 3s-Colorization \cite{yang2021skeleton}, and recently proposed 3s-AimCLR \cite{guo2022aimclr} into comparison. 
In these methods, test-time multimodal ensembling is performed and the results of using 1\% and 10\% labeled data are reported. 
We can find that our 3s-CMD still outperforms all of these methods after ensembling multiple skeleton modalities.

\subsection{Ablation study}
\setlength{\intextsep}{0pt}
\begin{wrapfigure}[17]{R}{0.4\textwidth}
	\centering
	\includegraphics[width=1.0\linewidth]{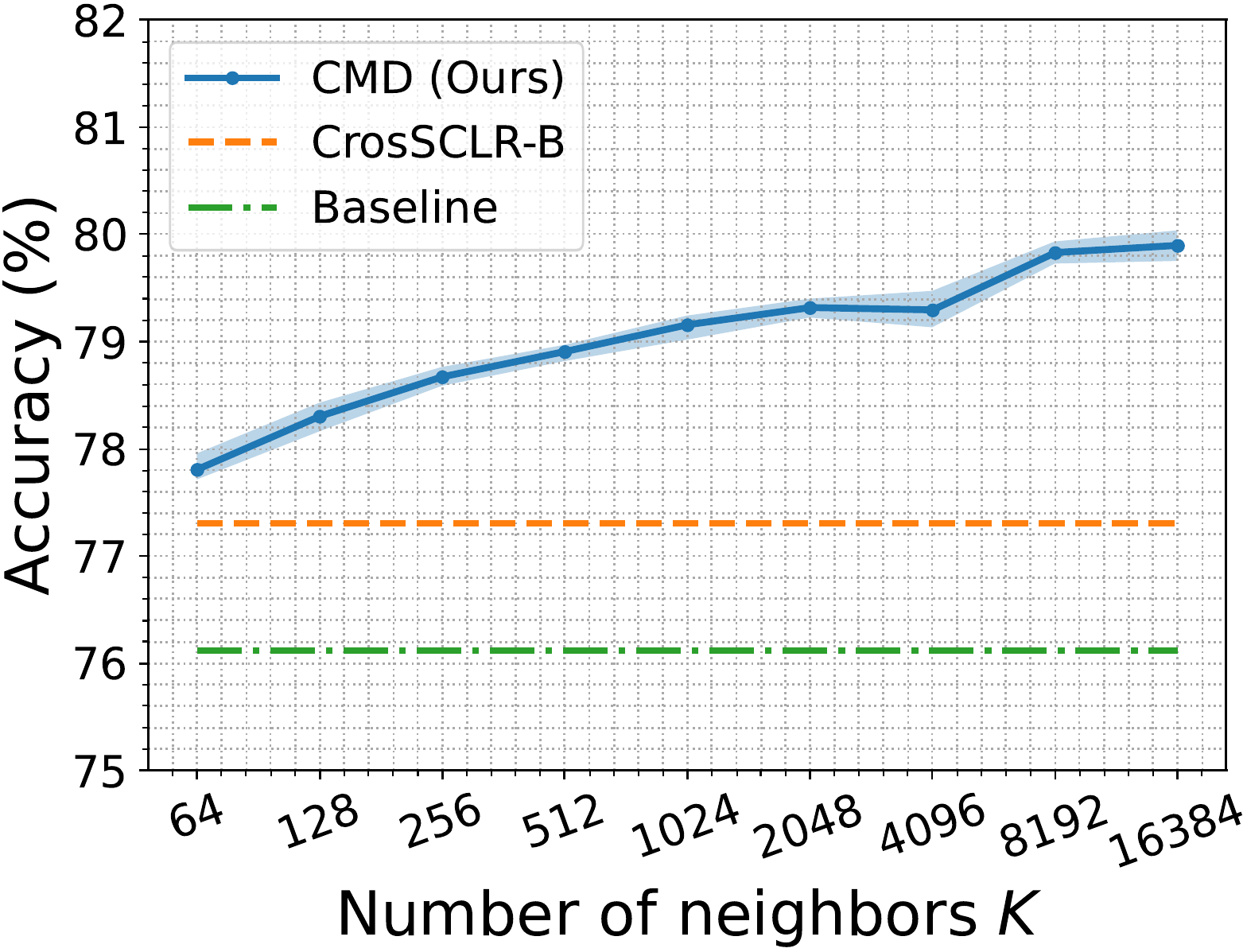}
	\caption{Ablative study of the number of neighbors $K$ in the cross-modal mutual distillation module. The performance is evaluated on the cross-subject protocol of the NTU-60 dataset.}\label{fig:topk} 
\end{wrapfigure}
To justify the effectiveness of the proposed cross-modal mutual distillation framework, we conduct several ablative experiments on the NTU-60 dataset according to the cross-subject protocol. More results can be found in the supplementary.

\parsection{Number of neighbors}
The number of nearest neighbors controls the abundance of contextual information used in the proposed cross-modal mutual distillation module. 
We test the performance of the learned representation with respect to different numbers of nearest neighbors $K$ under the linear evaluation protocol. 
As shown in Figure~\ref{fig:topk}, on the downstream classification task, the performance of the pre-trained encoder improves as $K$ increases.
When $K$ is large enough ($K \geq 8192$), continuing to increase its value hardly contributes to the performance. 
This is because the newly added neighbors are far away and contain little reference value for describing the distribution around the query sample.
In addition, we can also find that when the value of $K$ varies from 64 to 16384, the performance of our approach is consistently higher than that of CrosSCLR-B \cite{li20213d} and our baseline. 
This demonstrates the superiority and robustness of the proposed approach.

\begin{table}[t!]
	\centering
	\caption{Ablative experiments of modality selection and bidirectional distillation. The performance is evaluated on the NTU-60 dataset according to the cross-subject protocol. J, M, and B denote joint, motion, and bone modality respectively. The horizontal arrows indicate the direction of distillation.}\label{tab:abla_modality}
		\begin{tabular*}{1.0\linewidth}{p{0.27\linewidth}p{0.085\linewidth}<{\centering}p{0.085\linewidth}<{\centering}p{0.085\linewidth}<{\centering}p{0.075\linewidth}<{\centering}p{0.085\linewidth}<{\centering}p{0.085\linewidth}<{\centering}p{0.085\linewidth}<{\centering}p{0.075\linewidth}<{\centering}}
			\toprule
			\multirow{2}{*}{Modality \& Direction} & \multicolumn{4}{c}{\textbf{Linear Evaluation}} & \multicolumn{4}{c}{\textbf{KNN Evaluation}}\\
			\cmidrule(lr){2-5} \cmidrule(lr){6-9}
			& Bone  & Motion  & Joint & $\Delta$ & Bone  & Motion  & Joint & $\Delta$ \\
			\midrule
			Baseline 				& 74.4  & 73.1  & 76.1 &  & 62.0  & 56.8    & 63.4 & \\
			\midrule
			J $\leftarrow$ B     	& 74.4  & -     & 76.5 &  & 62.0  & -    & 64.3 & \\
			J $\rightleftarrows$ B 	& 76.6  & -     & 77.7 & $\uparrow$ 1.2 & 65.9  & -    & 66.5 & $\uparrow$ 2.2\\				
			\midrule
			J $\leftarrow$ M     	& -     & 73.1  & 78.9 &  & -  & 56.8    & 64.8 & \\
			J $\rightleftarrows$ M  & -     & \textbf{77.5}  & \textbf{79.8} & $\uparrow$ 0.9 & -  & 67.0  & 68.7 & $\uparrow$ 3.9\\
			\midrule
			J $\leftarrow$ M,
			J $\leftarrow$ B     	& 74.4 & 73.1   & 78.8 &  & 62.0  & 56.8    & 66.5 & \\
			J $\rightleftarrows$ M,
			J $\rightleftarrows$ B,
			M $\rightleftarrows$ B 	& \textbf{77.8} & 77.1  & 79.4 & $\uparrow$ 0.6 & \textbf{69.5} & \textbf{68.7} & \textbf{70.6} & $\uparrow$ 4.1\\
			\bottomrule
		\end{tabular*}
\end{table}

\parsection{Modality Selection} In our approach, we consider three kinds of skeleton modalities for self-supervised pre-training as in \cite{li20213d}. They are joint, motion, and bone, respectively. Our approach is capable of performing knowledge distillation between any two of the above modalities. As shown in Table~\ref{tab:abla_modality}, we report the performance of the representation obtained by pre-training with different combinations of skeleton modalities. Note that the joint modality is always preserved since it is used for evaluation. There are several observations as follows:
\textbf{i)} Cross-modal knowledge distillation helps to improve the performance of the representation in student modalities.
\textbf{ii)} Under the linear evaluation protocol, knowledge distillation between joint and motion achieves the optimal performance, exceeding the baseline by 3.7\%. \textbf{iii)} Under the KNN evaluation protocol, the learned representation shows the best results when all the three modalities are involved in knowledge distillation, which outperforms the baseline with an absolute improvement of 7.2\%.

\parsection{Bidirectional Distillation} In addition to modality selection, we also verify the effectiveness of bidirectional distillation. It enables the modalities involved in the distillation to interact with each other and progress together, forming a virtuous circle. In Table~\ref{tab:abla_modality}, the last column of each evaluation protocol reports the performance gain of bidirectional mutual distillation over unidirectional distillation in the joint modality. Results show that regardless of which skeleton modalities are used during pre-training, bidirectional mutual distillation further boosts the performance, especially under the KNN evaluation protocol.

\begin{figure}[t!]
	\centering
	\includegraphics[width=1.0\linewidth]{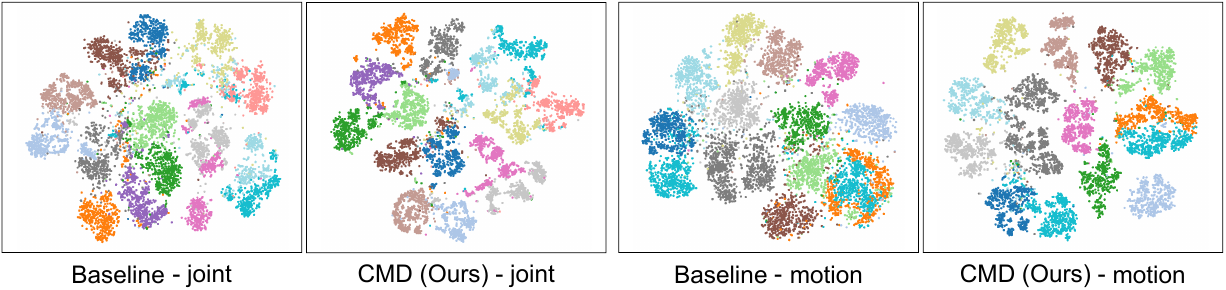}
	\caption{t-SNE \cite{van2008visualizing} visualization of feature embeddings. We sample 15 action classes from the NTU-60 dataset and visualize the features extracted by the proposed CMD and its baseline respectively. Compared with the baseline, CMD learns more compact and more discriminative representation in both joint and motion modalities.}\label{fig:tsne}
\end{figure}

\parsection{Qualitative Results} We visualize the learned representation of the proposed approach and compare it with that of the baseline. The t-SNE \cite{van2008visualizing} algorithm is adopted to reduce the dimensionality of the representation.  To obtain clearer results, we select only 1/4 of the categories in the NTU-60 dataset for visualization. The final results are illustrated in Figure~\ref{fig:tsne}. For both joint and motion modalities, the representation learned by our approach is more compactly clustered than those learned by the baseline in the feature space. This brings a stronger discrimination capability to the representation, explaining the stunning performance of our approach in Table~\ref{tab:sota_comp_knn}.

\section{Conclusion}
In this work, we presented a novel approach for self-supervised 3D action representation learning. It reformulates cross-modal reinforcement as a bidirectional knowledge distillation problem, where the pairwise similarities between embeddings are utilized to model the modality-specific knowledge. The carefully designed cross-modal mutual distillation module can be well integrated into the existing contrastive learning framework, thus avoiding additional computational overhead. We evaluate the learned representation on three 3D action recognition benchmarks with four widely adopted evaluation protocols. The proposed approach sets a series of new state-of-the-art records on all of them, demonstrating the effectiveness of the cross-modal mutual distillation.

~\\
\parsection{Acknowledgement} This work was supported by the National Natural Science Foundation of China under Contract U20A20183 and 62021001. It was also supported by the GPU cluster built by MCC Lab of Information Science and Technology Institution, USTC.

%
%
\bibliographystyle{splncs04}
\bibliography{egbib}
\end{document}